# A Computer Composes A Fabled Problem:
# Four Knights vs. Queen

Azlan Iqbal
College of Computer Science and Information Technology, Universiti Tenaga Nasional
Putrajaya Campus, Jalan IKRAM-UNITEN, 43000 Kajang, Selangor, Malaysia
azlan@uniten.edu.my

## ABSTRACT

We explain how the prototype automatic chess problem composer, Chesthetica, successfully composed a rare and interesting chess problem using the new Digital Synaptic Neural Substrate (DSNS) computational creativity approach. This problem represents a greater challenge from a creative standpoint because the checkmate is not always clear and the method of winning even less so. Creating a decisive chess problem of this type without the aid of an omniscient 7-piece endgame tablebase (and one that also abides by several chess composition conventions) would therefore be a challenge for most human players and composers working on their own. The fact that a small computer with relatively low processing power and memory was sufficient to compose such a problem using the DSNS approach in just 10 days is therefore noteworthy. In this report we document the event and result in some detail. It lends additional credence to the DSNS as a viable new approach in the field of computational creativity. In particular, in areas where human-like creativity is required for targeted or specific problems with no clear path to the solution.

## 1    INTRODUCTION

Chess problems or puzzles (also known as compositions or constructs) are essentially positions set up with a particular stipulation such as "*White to play and mate in three moves*". While seldom realistic they typically possess aesthetic or educational merit and the chess-playing community tends to find them enjoyable. Chesthetica is a prototype computer program that started off as an automatic evaluator of aesthetics or beauty in the game, specifically in three to five move direct mates and 'studies' (Iqbal et al., 2012). Studies typically include a more open-ended stipulation such as "*White to play and win*" and a longer solution. Chesthetica was then further developed into an automatic composer of chess problems using various computational creativity approaches (Iqbal, 2014; Iqbal et al., 2016). While the computer-generated problems thus far are not on the level of master human composers, this is due largely to a matter of specific requirements or conventions that master composers, esoterically, find aesthetically appealing. Compositions by human masters do not necessarily reflect the taste of the majority of chess-players and composers because they do not have extensive knowledge of compositions or share in their taste. This is not unlike how most people appreciate beauty in art but not necessarily the kind of art professional art critics and artists value highly. That does not mean the art that regular people appreciate is necessarily void of or lacking in aesthetics.

Regardless, Chesthetica's aesthetic evaluations have been shown experimentally to correlate positively and well with domain-competent human aesthetic assessment (Iqbal et al., 2012). This makes the automatic aesthetic assessment of thousands of chess problems viable and more consistent than if expert human composers were expected to do the job. Many of Chesthetica's compositions are also generally well-received in terms of having aesthetic or educational merit based on online community upvotes (Chesthetica Vidme, 2017). They are not the kind of problems that a 10 year old child who has been playing chess for a few years, for example, could easily be expected to compose from scratch. An artificial intelligence (AI) that outperforms children of that age in terms of creativity (even if just within a particular domain) is not insignificant. More importantly, the demonstrable mechanization of the creative process in this domain has far greater implications. In this article we will demonstrate how a particularly challenging type of chess problem was successfully composed by Chesthetica using the Digital Synaptic Neural Substrate (DSNS) AI approach. In section 2 we review Chesthetica's recent ability to compose custom mates and what led to the aforementioned challenging problem being chosen. In section 3 we explain the process through which the original four knights vs. queen problem was automatically composed. We discuss the implications of the work in section 4. The article concludes in section 5 with a brief summary and some direction for the future.

## 2 REVIEW

Starting with version 10.43 of Chesthetica, the program is able to compose custom mates as well (Iqbal, 2017). This means that a particular set of pieces can be specified and the program will try to compose a chess problem using only those pieces; no more and no less. In the aforementioned reference a chess problem by the renowned composer Pal Benko was used as an example. His problem featured a king, two bishops, a knight and two pawns versus a lone black king (KBBNPvk); see Fig. 1(a). The position in Forsyth-Edwards Notation or 'FEN' (Wikipedia, 2017), the stipulation and the main line of the solution are provided as well below the diagram. The primary appeal in this particular case was that all the pieces were arranged along a single file ('geometry'), not that White winning with such a large material advantage was difficult to imagine. The same set of pieces was input to Chesthetica and within a few hours several original chess problems using those pieces were composed. Notably, the one in Fig. 1(b).

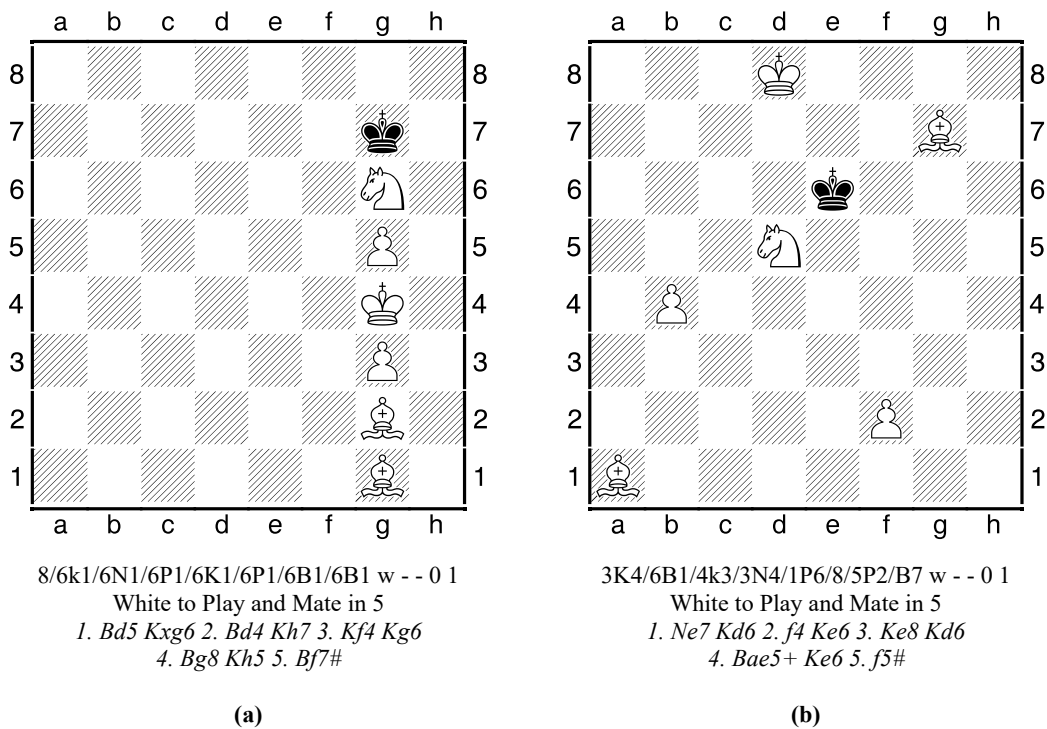

8/6k1/6N1/6P1/6K1/6P1/6B1/6B1 w - - 0 1
White to Play and Mate in 5
*1. Bd5 Kxg6 2. Bd4 Kh7 3. Kf4 Kg6
4. Bg8 Kh5 5. Bf7#*

(a)

3K4/6B1/4k3/3N4/1P6/8/5P2/B7 w - - 0 1
White to Play and Mate in 5
*1. Ne7 Kd6 2. f4 Ke6 3. Ke8 Kd6
4. Bae5+ Ke6 5. f5#*

(b)

**Figure 1**: Pal Benko's problem vs. Chesthetica's problem using the same pieces.

While Chesthetica's problem did not emphasize aesthetics in terms of geometry (which is just one of many principles of beauty in the game), it prompted a regular commenter at the site ("genem") to post the following at the bottom of the article.

*"I find the second puzzle particularly interesting, with its two white bishops both dark squares. It takes a specialized tactical eye to detect or build the kind of cages which surround Black's doomed king. Nonmate shot puzzles do Not train this type of tactical perception. The majority of mate shot puzzles don't really teach it either. Hmm, a gap in the body of chess literature? Thanks."* (Iqbal, 2017)

We considered that while the creativity of the program was demonstrable even with these pieces where there is a clear win for White, a more difficult set of pieces to force mate would be more illustrative of the program's creative capabilities. Inspired by the following quote from an earlier article, we decided to try the four knights vs. queen problem.

*"When I imagine Troitzky in his Siberian forest, surrounded by howling wolves, analyzing night after night whether king plus four knights can always beat king plus queen, that is great. That's what chess is all about, only you have to be a chess player to appreciate it. How can you explain to a non-chess player that within chess there is a little world of endgame studies...within which there is a microcosm made of utterly mad men analyzing four knights against queen"* (Havanur, 2016).

Endgame tablebases are databases that provide exhaustive analysis for chess positions with a certain number of pieces. At present, the limitation is 7 pieces or less constituting over 500 trillion positions *after* storage optimization (Lomonosov Tablebases, 2012) which includes the four knights vs. queen problem. We could not find any material explaining if all possible positions using this particular piece set had been tested to answer the question in the quote above. However, some testing using the "7-piece chess endgame training" Android app (Chess King, 2017) showed that most arbitrary positions set up using these pieces resulted in a forced mate in 30+ moves. However, the position was sometimes a draw given optimal play by both sides. So to answer the quote above, four knights cannot "always beat king plus queen". Endgame tablebases require that the position be set up in the first place in order for the determination of the result given optimal play to be made. They do not compose problems using particular piece sets, much less original studies and mates in 5, 4 or 3 moves which is what Chesthetica can do (and is furthermore not limited to 7 pieces). That kind of creativity has until recently been the domain of human chess problem composers.

Our search through various chess problem and puzzle books published within the last 50 years or so did not reveal any existing compositions using the four knights vs. queen piece set. It was perhaps to be expected since published chess problems tend to feature positions that use only the original piece set, i.e. no promoted pieces. It is yet another convention human composers are compelled to abide by in order to get published or compete in composition tournaments. The only one we managed to find was from a YouTube video showing a particularly long mate with these pieces (JMRWS YouTube, 2008). No composer or source was mentioned and not all the moves or even the length of the total solution are shown clearly in the video but the initial position appears as shown in Fig. 2(a). A strong chess engine such as Stockfish 8 would show the position as a draw but using the "7-piece chess endgame training" app with access to its omniscient tablebase of 7 pieces or less, it was shown to be a forced mate in 89 moves. Interestingly, the position was published to YouTube in 2008 which is about 4 years prior to the development of the first 7-piece endgame tablebase. So the solution played out in the video may not even be the most optimal ones but whoever composed it was correct that it was indeed a forced mate even though they could not have been certain of it.

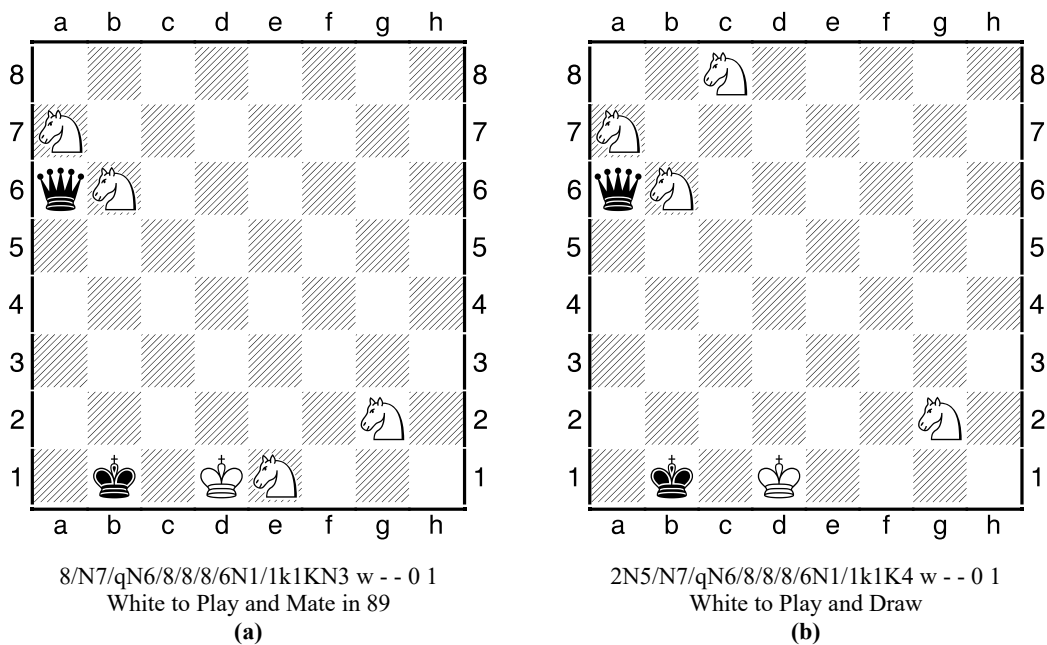

8/N7/qN6/8/8/6N1/1k1KN3 w - - 0 1  
White to Play and Mate in 89  
**(a)**

2N5/N7/qN6/8/8/6N1/1k1K4 w - - 0 1  
White to Play and Draw  
**(b)**

**Figure 2**: The only existing four knights vs. queen published won position found vs. a drawn variation of it.

To illustrate how important or delicate the initial position is using these pieces, see Fig. 2(b). Here, the knight is simply moved from the e1 to c8 square and the same app evaluates the position as a draw given optimal play by both sides. The actual moves and all the possible variations of these positions are not particularly relevant to the present discussion so they are not included here. In theory, there are easily tens of billions or more of legal positions with these pieces where none of the knights are under threat of clear capture (or 'en prise'). Most of them likely being wins or draws for White but not simple ones nonetheless. As shown above, mates can be as long as 89 moves. Part of the reason for the ambiguity of the win is derived from the idea that (using the typical material value of the chess pieces) three knights are equivalent to a queen which leaves just one knight against the king; a clear draw. Even two knights against a lone king is a clear draw. What makes the four knight vs.

queen piece combination even more challenging for humans is that, as mentioned earlier, it does not use the original piece set. Two pawns have been promoted to knights (unlikely in a real game) and the level of unfamiliarity is therefore higher than a position where there have been no promoted pieces, much less two underpromotions.

## 3   COMPOSING AN ORIGINAL FOUR KNIGHTS VS. QUEEN PROBLEM

A computer would have a considerably large domain or search space of positions to analyze systematically if composing a forced mate using these pieces was done using a brute force approach. A major feature of a good computational creativity approach is therefore the ability to reach meaningful or desirable locations in the search space without having to identify a particular path to it via brute force or other computationally-intensive methods. What we are dealing with here is essentially a form of the P versus NP problem, which is determining whether every problem where the solution can be verified relatively quickly by a computer can also be solved relatively quickly by a computer. So while a forced mate composition using four knights vs. queen can be solved relatively quickly using a game engine or 7-piece endgame tablebase, composing such a position automatically using a computer is not demonstrably as easy or straightforward, if possible at all. Given that the four knights vs. queen piece set is relatively difficult to compose with (even for humans), it served as a worthy challenge to Chesthetica and the DSNS technology it incorporates. The DSNS essentially uses feature information from various domains such as chess move sequences, images and music, in combination with stochastic methods (a kind of randomness) to help generate objects in any of the original domains.

For instance, forced mate move sequences taken from tournament games in combination with photographs of people could be used to generate original chess problems. This is not unlike the creative process as it occurs, often unpredictably, in humans. Some people call it 'inspiration' and it can be due to being exposed to various types of stimuli at the right time or in the right mix. Readers interested to learn more about how the DSNS is used by Chesthetica to compose chess problems should read (Iqbal et al., 2016 and Iqbal, 2016). Details regarding the workings of the aesthetics model that was developed prior to the DSNS can be found in (Iqbal et al., 2012). It is important to emphasize that the DSNS has nothing to do with artificial neural networks (ANNs) or even machine learning, which are common misconceptions. It is an entirely new computationally creative approach or method. Besides, creativity as far as we know is not something that someone typically 'learns'. It simply 'arises' or pops into mind at unpredictable times and for no reason apparent to our consciousness. Fig. 3 shows the interface to Chesthetica (presently in version 10.53) and how the piece requirement was set up for attempted composition. The position shown in the main window (right) is not the one that was successfully composed but rather taken in the midst of the general composing process as it would have run (it happens to be a draw with White to move). A video of said process in motion, using an earlier version of the program, is available in (Chesthetica YouTube, 2016).

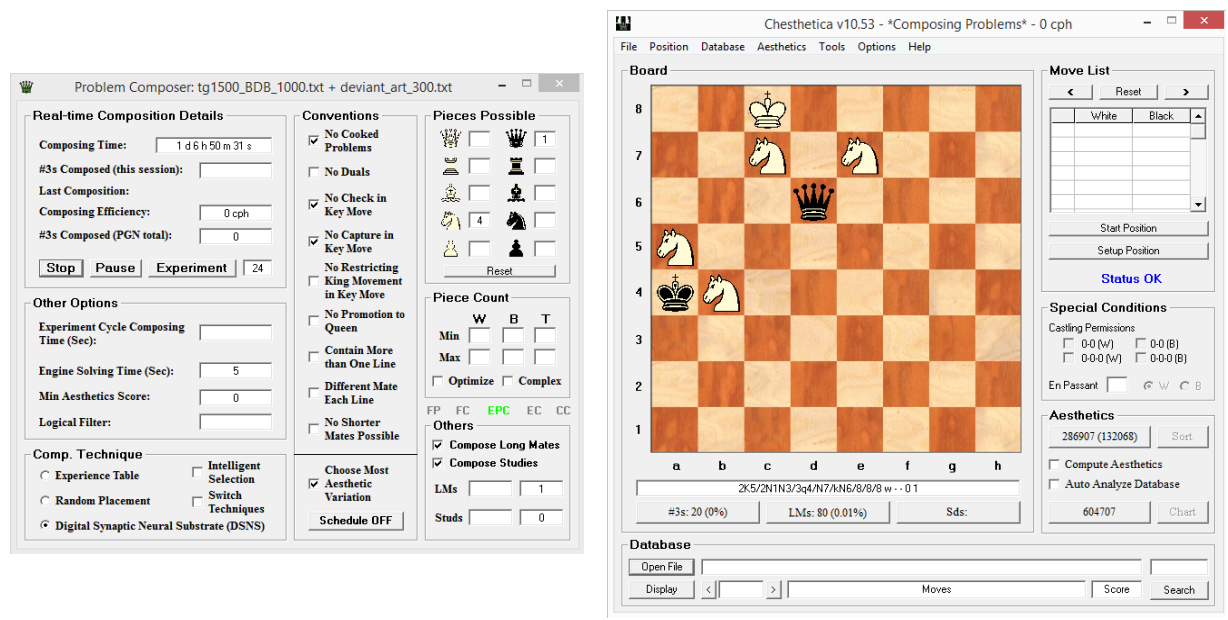

**Figure 3**: Chesthetica set up to compose the four knights vs. queen problem.

As is evident in the 'composer' window (left), Chesthetica can now even compose studies using particular piece sets (in addition to forced three, four and five move mates) depending on user requirements. Interested readers may sample the variety and range of these compositions which are uploaded online daily (Enemark, 2015; Chesthetica YouTube, 2017). Since we were not sure of the length or type of problem that could be successfully composed with the aforementioned pieces, we selected all the four problem types. Three conventions were also applied, namely, *no cooked problems*, *no check in key move* and *no capture in key move*. These can be seen as checked in the composer window. A 'cooked' problem, by the way, is one where there is an additional key (i.e. first) move not intended by the composer. Applying these conventions helps ensure that the composition is not simply one where the key move was say, forking the king and queen and then winning the queen. That would not be particularly interesting or having aesthetic merit even though composing something like that would not be all that straightforward or easy either, even for a human.

### 3.1 The First Failed Attempt

An effort to compose this particular rare type of chess problem using Chesthetica had actually started somewhat earlier on, i.e. before the successful attempt. In the earlier attempt, the DSNS approach used tournament chess game mate sequences between relatively weak players in combination with photographs of people (Iqbal, 2016) in order to compose the four knights vs. queen problem because this particular combination had shown the best results thus far in composing chess problems, in general. Curious readers should go through (Iqbal et. al, 2016 and Iqbal, 2016) to understand why. The computer we used in the failed composing attempt was a small Asus 'Transbook' with an Intel(R) Atom™ CPU Z3775 @ 1.46GHz, running Microsoft Windows v8.1 32-bit with 2 GB of RAM. More specifically, Chesthetica was running simultaneously on two user accounts on this machine and only one account was working to compose a viable four knights vs. queen chess problem. So the available processing power and memory of this limited machine was, in fact, theoretically halved.

We cannot be certain if more processing power necessarily produces better results given a creative computing approach (the DSNS or any other method) but it seems to make sense that it would or should. Similarly, having many instances of the program running on many different machines should increase the chances of success as well. This is not unlike many different human research groups working on the same problem. The chances of success should be higher than if only one group was working on the problem. The instance of Chesthetica in question, just like the other one running on the other user account (but composing other problems), was running for 24 hours a day, 7 days a week for approximately a month but produced no results. There was not a single, viable composition of the type required (the specifications as shown in Fig. 3) produced. This is not to say that no result would ever have been produced but it just so happened that a new set of images was ready that could be applied to the DSNS composing process and we decided to try it instead.

### 3.2 The Second Successful Attempt

The set of images used in the failed attempt above (300 images of people from the Bigstock online photo archive) did not appear to produce any result with regard to the four knights vs. queen problem. Samples of these pictures can be seen in Fig. 1 of (Iqbal, 2016). How they were used together with the tournament chess game mate sequences is explained in section 2 of the same reference. The new set of images in the second attempt was not from Bigstock but the DeviantArt website, an online art gallery and community (DeviantArt, 2017). The site features fascinating and often provocative art by people from all over the world where users can select 'favorites'. The collection of 300 images from this site was our own personal selection from many months of casual browsing. It was based on personal taste and performed specifically with the intention of creating a new, unrushed image set for use with the DSNS. These images were not limited to photographs of people but also included, among other things, photographs and drawings of people and places (both real and imagined). A visit to the DeviantArt website will give the reader a better idea of what kind of images they have. Appendix A shows four sample images (from the 300) that were used. The argument could be made that the greater variety and range of imagery in this set possesses more 'creative content' for the DSNS to draw upon. However, this is both difficult to measure and demonstrate experimentally. It was nevertheless a contributing general idea behind the use of this new image set.

#### 3.2.1 The Original Four Knights vs. Queen Problem Composed

Approximately 10 days of processing using the same computational setup as explained in section 3.1 but applying the new DeviantArt image set (as explained in section 3.2 above) instead of the Bigstock set proved successful. It resulted in the creation of a forced mate in 5 moves that fulfilled all the specified criteria as also shown in the composer window (left) of Fig. 3. The forced mate in 5 can be seen in Fig. 4. Provided below the diagram is the position in FEN, the version of Chesthetica that composed it, the stipulation, the location, and the precise date and time of composition.

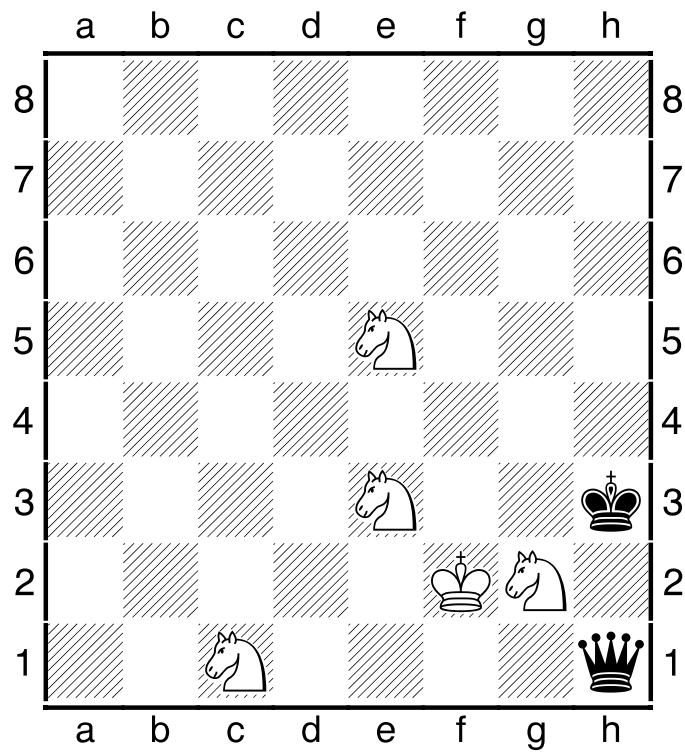

8/8/8/4N3/8/4N2k/5KN1/2N4q w - - 0 1
Chesthetica v10.53
White to Play and Mate in 5
Selangor, Malaysia
2017.8.23 4:51:51 AM

**Figure 4**: The original four knights vs. queen chess problem composed autonomously by Chesthetica.

The position was confirmed to indeed be a forced mate in 5 using both the Stockfish 8 engine and the "7-piece chess endgame training" app. Incidentally and after the fact, Chesthetica evaluated this composition as scoring 3.086 aesthetically which is rather good and definitely within the class of what most human players with domain competence would consider a chess problem (see footnote 6 in Iqbal et al., 2012). Chesthetica's aesthetic evaluation capabilities actually do not factor into the composition process unless specified using the "Min Aesthetics Score" (see Fig. 3) and for this particular task that additional constraint was not applied. Even so, it is notable that the result was still of fairly good aesthetic merit. The main line of the solution is as follows: 1. Ne2 Qxg2+ 2. Nxg2 Kh2 3. Ngf4 Kh1 4. Ng3+ Kh2 5. Nf3#. Chesthetica was set to choose the most aesthetic variation based on its internal model as the main line, just as a human composer would typically do. A graphical representation of the moves of this main line in the form of several diagrams are in Appendix B for the benefit of those who have trouble visualizing the moves.

The relatively complete analysis (i.e. including other variations or lines of play) also happens to be rather compact as is presented in Fig. 5. By 'compact' we mean that there is not an enormous number of variations to consider. For some problems of the same length (or even shorter) there can be as many as 20,000 or more lines of play even though there is only one key move. While having many variations is not always a major impediment it does make digesting the problem more difficult for human solvers. It also likely makes the problem somewhat less aesthetically pleasing from a psychological standpoint (Margulies, 1977). Another feature of the key move (Ne2) is also that it is a 'quiet' move, meaning that it does not check the king or capture a piece. Key moves that are *not* quiet are considered too obvious and frowned upon in problem composition. These composition conventions were actually specified in advance as filters (see Fig. 3) so Chesthetica had to abide by them. The more filters one applies, the less likely or quickly the program is able to produce a result. Similarly, the more requirements or constraints one places on a human composer in terms of conventions to follow, the less likely or quickly he is able to compose the problem.

| | | | | | | | |
|---|---|---|---|---|---|---|---|
| Ne2 | Qg1+ | Nxg1+ | Kh2 | Nf1+ | Kh1 | Ng3+ | Kh2 | Nef3# |
| | | | | | | | | Ng4# |
| | | | | Nf5 | Kh1 | Ng3+ | Kh2 | Nef3# |
| | | | | | | | | Ng4# |
| | Qf1+ | Kxf1 | Kh2 | Nf3+ | Kh3 | Ngf4# |
| | | | | | Kh1 | Ng3# |
| | Qe1+ | Kxe1 | Kh2 | Nf3+ | Kh3 | Ngf4# |
| | | | | | Kh1 | Ng3# |
| | Qxg2+ | Nxg2 | Kh2 | Ngf4 | Kh1 | Ng3+ | Kh2 | Nf3# |
| | | | | | | | | Ng4# |
| | Qd1 | Ngf4+ | Kh4 | Nf3# |
| | | | Kh2 | Nf3+ | Kh1 | Ng3# |
| | | | | N5g4+ | Kh1 | Ng3# |
| | | | | N3g4+ | Kh1 | Ng3# |
| | Qc1 | Ngf4+ | Kh4 | Nf3# |
| | | | Kh2 | Nf3+ | Kh1 | Ng3# |
| | | | | N5g4+ | Kh1 | Ng3# |
| | | | | N3g4+ | Kh1 | Ng3# |
| | Qb1 | Ngf4+ | Kh4 | Nf3# |
| | | | Kh2 | Nf3+ | Kh1 | Ng3# |
| | | | | N5g4+ | Kh1 | Ng3# |
| | | | | N3g4+ | Kh1 | Ng3# |
| | Qa1 | Ngf4+ | Kh4 | Nf3# |
| | | | Kh2 | Nf3+ | Kh1 | Ng3# |
| | | | | N5g4+ | Kh1 | Ng3# |
| | | | | N3g4+ | Kh1 | Ng3# |
| | Qh2 | Ng1+ | Qxg1+ | Kxg1 | Kg3 | Nf5+ | Kh3 | Nf4# |
| | Kh2 | Nf3+ | Kh3 | Ngf4# |
| | | | | Nef4# |
| | | N5g4+ | Kh3 | Nef4# |
| | | N3g4+ | Kh3 | Nef4# |

**Figure 5**: Analysis of the original four knights vs. queen chess problem composed autonomously by Chesthetica.

One could argue that the key move of 1. Ne2 is simply moving the knight out of harm's way, i.e. from being captured by the queen yet playing 1. Nd3 achieves the same goal. Interestingly, in this position, 1. Nd3 wins as well but in 9 moves rather than 5 so it is not the most optimal key move and does not satisfy the stipulation. Regardless, having a position that is both a mate in 5 and a mate in 9 is more challenging to compose than either one. This was not something prespecified to the program. After 1. Ne2, Black has a variety of choices as seen in the second column of Fig. 5. Note that suboptimal play by Black could lead to even shorter mates. In the main line, Black replies with 1. … Qxg2+. This technically forms a three knights vs. queen mate in 4 problem but not a particularly attractive one because White's key move in this position is rather obvious. Should White reply incorrectly with 2. Ke1, for instance, that would be a drawn position featuring three knights vs. queen given optimal play by both sides. After 2. Nxg2 Kh2, we have a forced mate in 3 position featuring three knights vs. lone king. This is a reasonable chess problem in itself because the key move in this position would be the quiet 3. Ngf4, even though 3. Ne3 (and several other possibilities) also checkmates but in 5 moves.

After 3. … Kh1, White can also win in three moves by playing, for example, 4. Ke1 instead of 4. Ng3+ as in the main line so optimal play is still matters even at this stage of the problem. It is nice to win but realizing you could have done so sooner often takes something away from the aesthetics of it all. Another notable trait of the main line is the geometry in move 4, see Appendix B(h) and (i). We saw a similar emphasis on geometry in Pal Benko's problem in Fig. 1(a). Finally, even the checkmate position has a triangular geometry about it. None of these were prespecified to Chesthetica either. So there are many aspects to this composition that makes it unlikely to have been the product of random chance. The DSNS process and perhaps even the new DeviantArt image set likely aided somehow in the computationally creative process. Simply throwing these pieces randomly on the board (even millions of times over) would likely not have formed any kind of checkmate whatsoever given the number of possibilities, much less a forced mate in 5 with aesthetic qualities and features as explained above. Indeed, the DSNS has been tested in the past and outperformed not only random chance but also the 'experience table' composing approach that was developed prior to it (Iqbal et al., 2016).

## 4      DISCUSSION

The automatic composition of an original four knights vs. queen problem may seem trivial to master human chess problem composers who perhaps feel threatened by new technological developments with regard to computer chess. A similar thing happened with chess-*playing* programs that started to appear in the 1950s. They were essentially mocked for decades as being unpromising and never able to play at the master level (Dreyfus, 1979). This changed in 1997 when the world chess champion Garry Kasparov lost to IBM's Deep Blue in a six-game match (Pandolfini, 1997). Since then, even the significantly more complex Japanese game of go has been conquered by machine intelligence (Byford, 2017). However, creative tasks are somewhat different from game-playing where the objective is much clearer and easier to judge. With regard to Chesthetica, Frederic Friedel (co-founder of the popular chess website, 'ChessBase') says:

*"On the downside it is slightly depressing to consider that the program would then be able to generate hundreds of valuable compositions per day, producing in a week what it took a talented human composer a lifetime to achieve. In any case I wish Dr. Iqbal success in his endeavour. It is pioneering work that will be seen, in similar vein, in a large number of fields of human activity. This is the way AI will develop and progress."* (Friedel, 2017)

It is not clear whether Chesthetica will one day routinely compose the type of chess problems that human master composers (a vanishingly small subset of the global chess community) tend to prefer even though it does not seem like a technological impossibility to us. Even so, the goal here is unclear. There is no 'world champion' composer to beat that would prove decisively which compositions are objectively the most beautiful. There is also no company like IBM interested and willing to invest tens of millions of dollars and years of additional research into the development of specialized hardware and software tailored specifically to this task. As it stands, Chesthetica is indeed able to compose chess problems of the kind that clearly illustrate creativity. This is because had a 10 year old child composed the kinds of problems it can compose, we would surely attribute creativity to them. Now, with the customized composition of a four knights vs. queen problem, this idea of creativity is even harder to deny.

A non-trivial and (unlike before) specific set of requirements (i.e. a 'creative problem') was given to the program and a solution or answer was indeed produced within a reasonable period of time. Furthermore, the solution was unpredictable in terms of what, exactly, it would look like and when it would occur. These are known characteristics of creativity even in humans. This is also what differentiates traditional AI approaches from the DSNS computational creativity approach. We do not have all the answers with regard to why the DSNS appears to work in computational creativity tasks (at least within the domain of chess problem composition) or even how to manipulate it to maximize output and quality even though previous work (Iqbal, 2016) indicates that larger and higher quality images can improve results. Given the present report, it would seem that a set of images arguably infused with more creative content can improve performance of the DSNS as well. One idea from outside the field suggests that different objects ultimately or at some point share an underlying physics which facilitates 'translation' from one form or domain to another (Lin, Tegmark and Rolnick, 2017).

While it may be important that AI be able to 'explain itself' in order to aid our understanding of AI decision-making, when it comes to a creative task we would not expect even a human to be able to explain the process because the human is simply unaware of how or why certain things pop into mind. So while the DSNS process can be understood (Iqbal et al., 2016), the 'precise steps' that led to the creation of the present four knights vs. queen composition would not add much to the discussion. AI over the last 60 years has had periods of both success and 'winters'. Even presently with the field enjoying some success and significant media coverage, there are skeptics who doubt the promise of AI (Intelligence Squared, 2016). Among many other issues, they believe that it has been reduced to mere statistical analyses of massive amounts of data and has little to do with the kind of intelligence and creativity humans possess. This is precisely why fundamentally new approaches such as the DSNS which address creativity specifically are important and worth pursuing, as opposed to 'tried-and-true' (old) AI methods or even newer variations of them.

There are also those that perhaps overestimate the potential and consequences of AI, believing it will exceed human intelligence and possibly threaten the species before too long (Bostrom, 2016; Harris, 2016). Most of these concerns are perhaps unfounded because they fail to take into account the 'experimental bottleneck' or the fact that new knowledge acquisition is not simply a matter of neuronal processing speed but rather requires carefully and creatively designed experiments that consume money, energy and especially time. Not to mention flexible limbs and opposable thumbs, in most cases.

## 5    CONCLUSIONS

The DSNS is a relatively new computational creativity approach. It has been incorporated into a computer program called Chesthetica and since around August 2014 used to compose chess problems of various types which have been published online (automatic composition using earlier methods started around July 2010). Newer features are periodically included into the program to expand its capabilities. Notably among the most recent was the ability to compose custom mates or puzzles (typically forced checkmate in three, four or five moves). This meant that a more specific 'creative task' could be specified to the program, thereby increasing the value of its output. For example, not just 'mate in 3' but 'mate in 3 using three knights against lone king'. No longer would we simply have to accept whatever piece count and combinations of pieces the program decided to use to compose a problem (which is still useful and demonstrate creativity) but a specific piece set could now also be specified and the program had to work with what it was given.

With the successful composition of a rare chess problem type, i.e. four knights vs. queen, the capabilities of the DSNS approach as a viable technology capable of being incorporated into a computationally creative software is illustrated even further. While the potential of the DSNS in areas other than chess problem composition remain to be demonstrated, there appears to be little left to achieve in the domain of chess problem composition. The goal of composing the types of problems enjoined by master problem composers is not something worth pursuing without sufficient financial backing and support. It does not add any more to the AI literature than an extremely specialized task in a very narrow domain would. Besides, the vast majority of the world's chess playing community have little to no patience and appreciation for the musings of master composers. To claim that only chess compositions of the type composed by master composers should be deemed beautiful or aesthetic is tantamount to claiming that all art is worthless except the works of artists like Picasso or van Gogh.

Most people would actually prefer to have other types of art (by 'unknown' artists) decorating their homes and offices. Given the social media explosion in recent years and feedback with regard to what Chesthetica has produced over the years, we are satisfied that there is indeed an audience which appreciates the level of aesthetics and creativity it demonstrates even now. Future work will therefore likely include more compositions being generated (since there appears to be no limit to the creative output and range of the DSNS in this domain) and uploaded to sites such as YouTube, Vidme and Minds. In addition, perhaps more publications and books featuring some of the choicest compositions by Chesthetica, proving that the program is able to create value and wealth as well. Our hope, of course, is that the DSNS approach will eventually be picked up by researchers in other areas (perhaps with more pressing problems to deal with) and that it will prove as effective as it is in the present domain, if not more so.

**APPENDIX A:** Sample DeviantArt Images Used in the Successful DSNS Composing Process

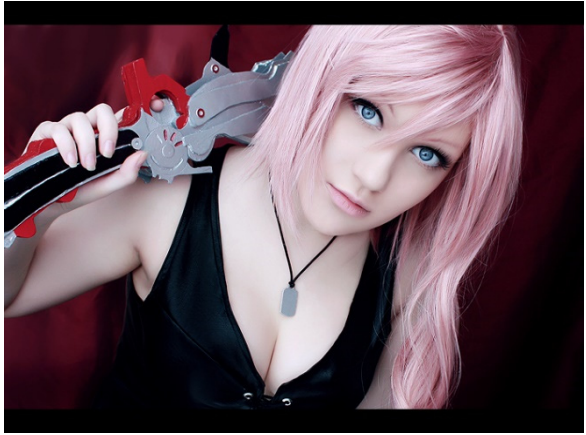

(a) Lightning - Aya
(https://katy-angel.deviantart.com/art/Lightning-Aya-192153015)

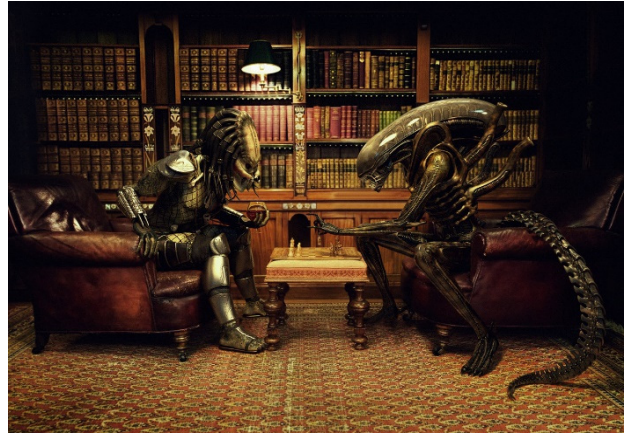

(b) Alien Vs. Predator: Chess
(https://xidon.deviantart.com/art/Alien-Vs-Predator-Chess-165322037)

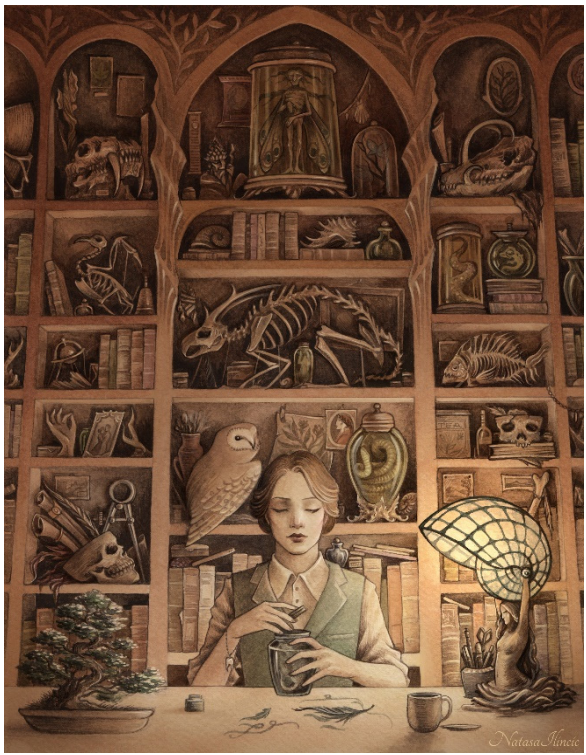

(c) Magpie
(https://natasailincic.deviantart.com/art/Magpie-651371060)

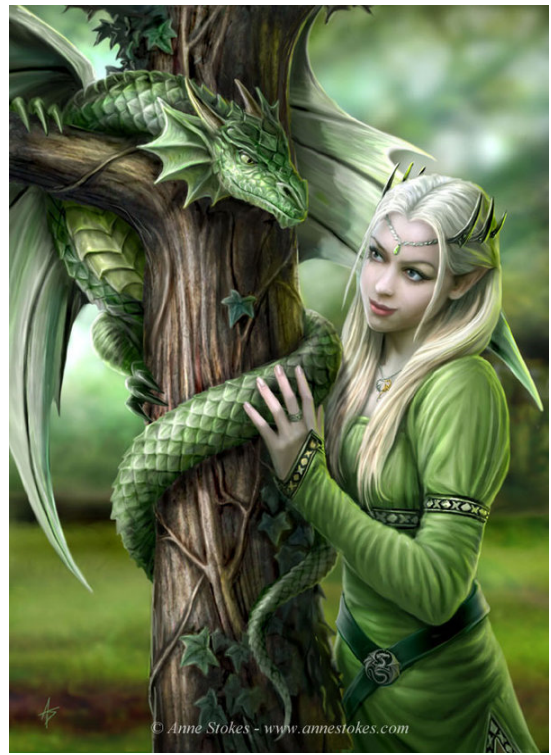

(d) Kindred Spirits
(https://ironshod.deviantart.com/art/Kindred-spirits-113299119)

**APPENDIX B:** Main Line of the Original Four Knights vs. Queen Chess Problem Composed by Chesthetica

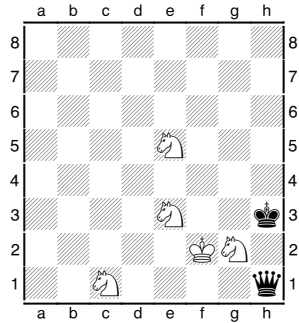
(a) White to play and mate in 5

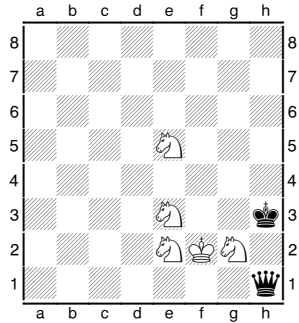
(b) 1. Ne2

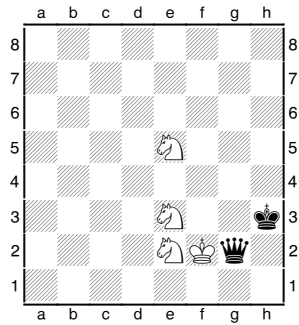
(c) 1. … Qxg2

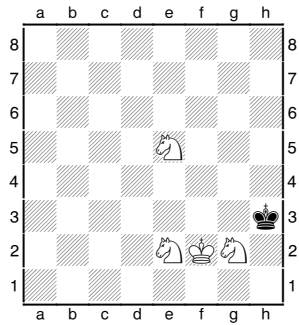
(d) 2. Nxg2

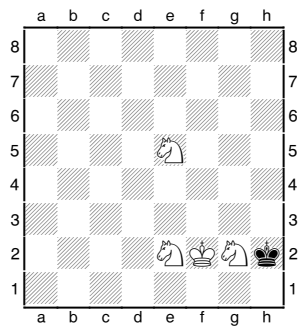
(e) 2. … Kh2

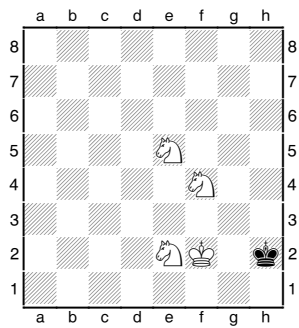
(f) 3. Ngf4

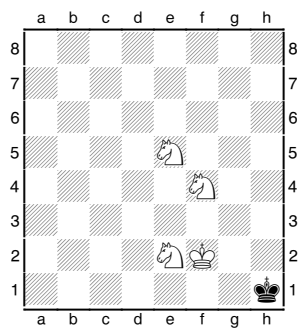
(g) 3. … Kh1

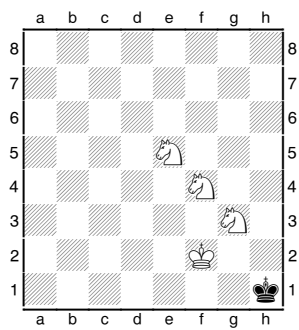
(h) 4. Ng3+

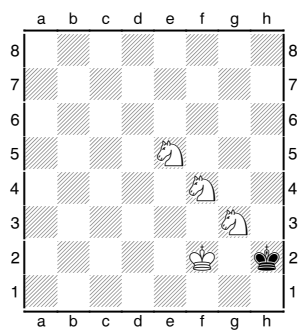
(i) 4. … Kh2

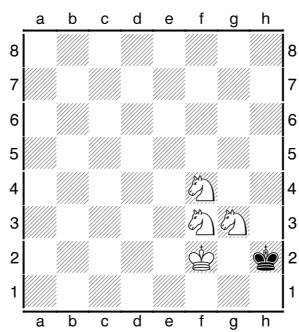
(j) 5. Nf3#